\documentclass[10pt,twocolumn,letterpaper]{article}

\usepackage{cvpr}
\usepackage{times}
\usepackage{epsfig}
\usepackage{graphicx}
\usepackage{amsmath}
\usepackage{amssymb}

\usepackage[pagebackref=true,breaklinks=true,letterpaper=true,colorlinks,bookmarks=false]{hyperref}

\cvprfinalcopy 

\def\cvprPaperID{9483} 

\ifcvprfinal\fi
\begin{document}

\title{\textit{Show, Edit and Tell}: \protect\\ A Framework for Editing Image Captions}

\author{Fawaz Sammani$^{1}$, Luke Melas-Kyriazi$^{2}$\\
$^{1}$Multimedia University, $^{2}$Harvard University\\
{\tt\small fawaz.sammani@aol.com, \tt\small lmelaskyriazi@college.harvard.edu}
}

\maketitle

\begin{abstract}

Most image captioning frameworks generate captions directly from images, learning a mapping from visual features to natural language. However, editing existing captions can be easier than generating new ones from scratch. Intuitively, when editing captions, a model is not required to learn information that is already present in the caption (i.e. sentence structure), enabling it to focus on fixing details (e.g. replacing repetitive words). This paper proposes a novel approach to image captioning based on iterative adaptive refinement of an existing caption. Specifically, our caption-editing model consisting of two sub-modules: (1) \textit{EditNet}, a language module with an adaptive copy mechanism (\textit{Copy-LSTM}) and a Selective Copy Memory Attention mechanism (\textit{SCMA}), and (2) \textit{DCNet}, an LSTM-based denoising auto-encoder. These components enable our model to directly copy from and modify existing captions. Experiments demonstrate that our new approach achieves state-of-art performance on the MS COCO dataset both with and without sequence-level training. Code can be found at \href{https://github.com/fawazsammani/show-edit-tell}{https://github.com/fawazsammani/show-edit-tell}. 

\end{abstract}

\section{Introduction}
Image captioning is the task of producing a natural language description of a visual scene. As one of the prototypical examples of multimodal learning, image captioning combines techniques from computer vision (e.g. recognizing salient objects in an image), with those from natural language processing (e.g. generating coherent sentences describing these objects). Applications of image captioning include content-based image retrieval \cite{ordonez2016retrieval} and assisting the visually impaired by converting visual signals into text, which can then be transformed to speech using text-to-speech technologies \cite{macleod2017understanding}.

\begin{figure}
    \centering
    \includegraphics[width=0.5\textwidth]{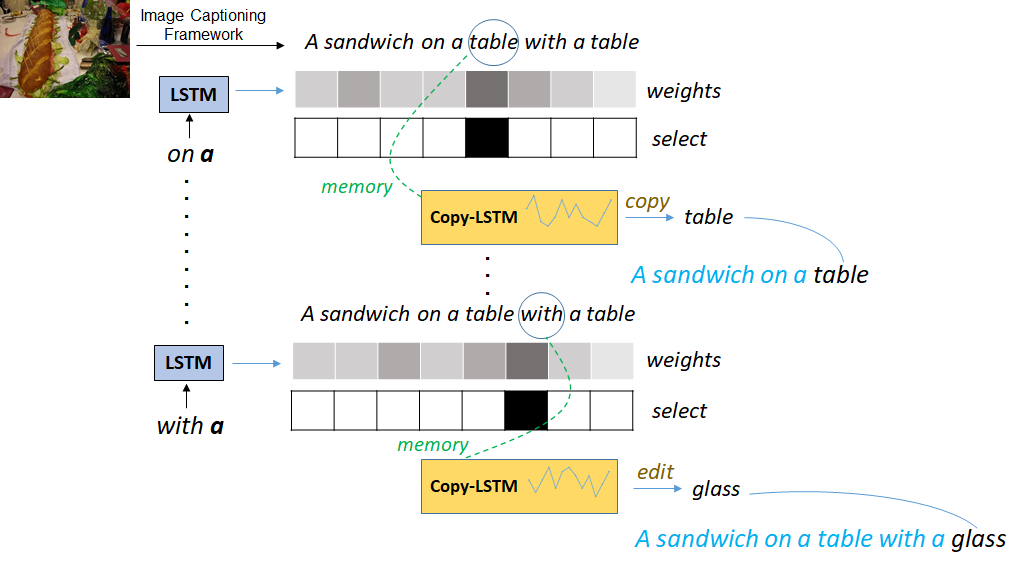}
    \caption{Our model learns how to edit existing image captions. At each decoding step, attention weights (grey) are generated; these correspond to the importance of each word in the existing caption for the word currently being generated in the new caption. Using a selective copy memory attention (SCMA) mechanism, we select the word with the highest probability and directly copy its corresponding LSTM memory state to our language LSTM (Copy-LSTM). That is, rather than learning to copy words directly from the input caption, we learn whether to copy the hidden states corresponding to these words. We then generate our new caption from this (possibly copied) hidden state. Best viewed in color.}
    \label{demo}
\end{figure}

Over the past five years, neural encoder-decoder systems have gained immense popularity in the field of image captioning due to their superior performance compared to previous image processing-based techniques. 
The current state-of-art image captioning models are composed of a CNN encoder, an LSTM (or Transformer) decoder, and one or more attention mechanisms. The input image is first encoded by a CNN into a set of feature vectors, each of which captures semantic information about an image region, and these feature vectors are decoded using an LSTM-based or Transformer-based network, which generates words sequentially. Attention mechanisms enable the decoding process to ``focus'' on particular image regions during generation, and the specific formulation of these mechanisms has been the center of much research \cite{Huang2019AttentionOA,Anderson2017BottomUpAT,Xu2015ShowAA,Lu2017Adaptive,You2016SemanticA}. 

High-quality captions consist of two elements: coherent natural language sentences (i.e. sentence/caption structure), and visually-grounded content (i.e. accurate details). Current image captioning models learn a ground-up mapping from image features to full captions, hoping to capture both elements simultaneously. Examining the outputs of prior image captioning models \cite{Xu2015ShowAA, Anderson2017BottomUpAT, Yao2018ExploringVR}, we observe that models learn global sentence/caption structure exceptionally well, but they often produce incorrect, inconsistent, or repetetive content.

Motivated by this observation, in recent months, researchers have begun to consider the problem of editing inputs independent from the problem of generating inputs \cite{Hashimoto2018ARF,Sammani2019LookAM}. Intuitively, editing should be easier than generating from scratch, because a caption-editing model can focus on visually-grounded details rather than on caption structure \cite{Sammani2019LookAM}. For example, consider Figure \ref{demo}: A state-of-art image captioning framework \cite{Huang2019AttentionOA} outputs ``A sandwich on a table with a table.'' The network produces a sensible sentence structure for this particular image (``A \_\_ on a \_\_ with a \_\_'') but fails to properly fill in the nouns, repeating the main object in the image (``table''). A caption-editing model, applied to this caption, should be able to recognize this error (the noun repetition) and modify the caption to read ``A sandwich on a table with a glass of wine'' or perhaps simply ``A sandwich on a table''. 

We propose a novel approach to image captioning based on iterative adaptive refinement of an existing caption rather than from-scratch caption generation. At each decoding step of the caption editing process, a word from the existing caption is selected and its corresponding memory state is copied into the internal structure of the LSTM (the \textit{Copy-LSTM}). This Copy-LSTM includes a separate selective copy attention mechanism (SCMA), enabling it to further edit or copy the existing word into the final output caption. For example, in Figure \ref{demo}, our model chooses to copy the first instance of the word ``table'' and edit the second instance to ``glass''. Ultimately, our model produces: ``A sandwich on a table with a glass of wine''. 

In summary, our contributions are as follows: 

\begin{itemize}
\item We propose EditNet, a framework for editing existing image captions that consists of a Copy-LSTM equipped with a Selective Copy Memory Attention (SCMA) mechanism. Alongside EditNet, we propose DCNet, a denoising auto-encoder that learns to denoise previous captions. We optimize DCNet with a novel objective function (MSE between hidden states), finding it to be a simple and effective way to improve the performance of our decoder.

\item Our method achieves a new state-of-the-art performance on MS COCO dataset.

\item We present an ablation analysis of the components of our model, demonstrating that each aspect contributes non-trivially to our model's final performance.

\end{itemize}
 
\begin{figure*}
    \centering
    \includegraphics[width=\textwidth,height=\textheight,keepaspectratio]{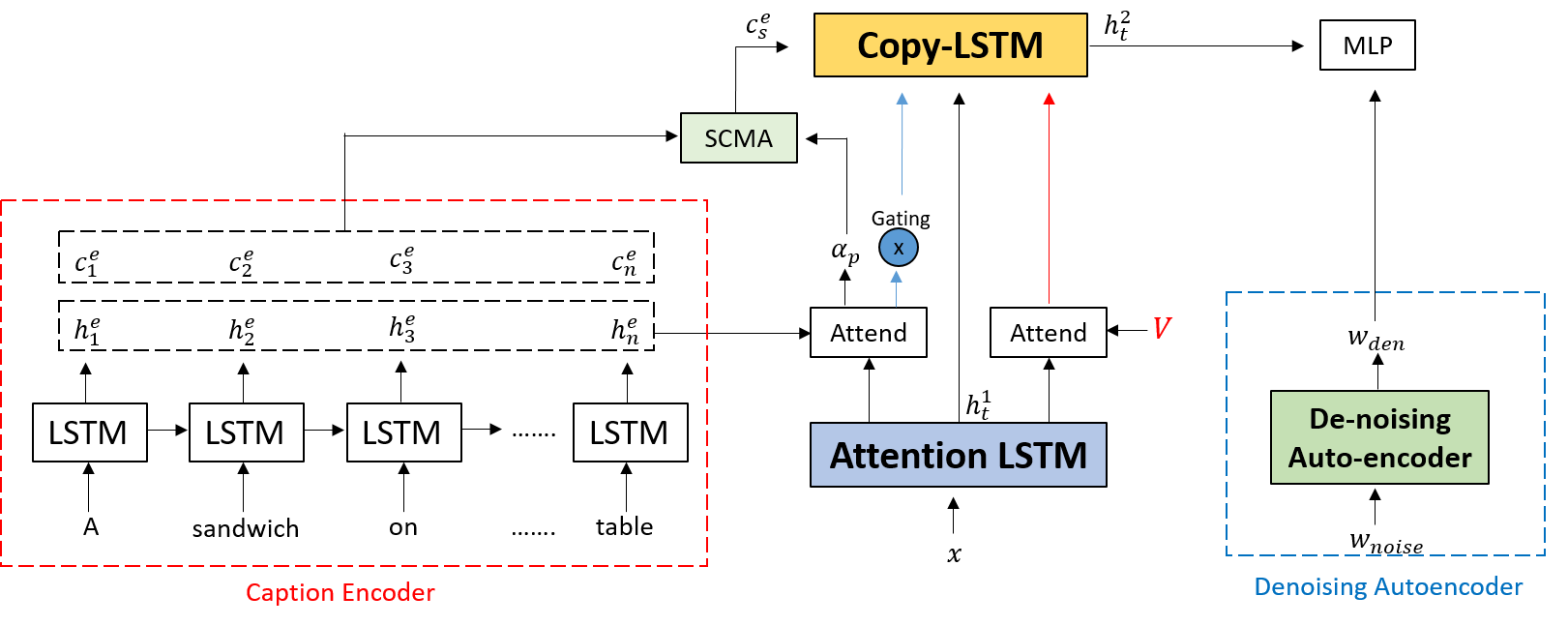}
    \caption{Our proposed model which includes two submodules: On the left, EditNet along with its decoder (middle) is shown. For EditNet, the existing caption is first encoded using a uni-directional LSTM, where each LSTM cell outputs a word representation (the hidden state $h_{t}^{e}$ and a memory state $c_{t}^{e}$). The hidden states are used to calculate attention weights which are then passed to the SCMA mechanism along with the memory states. The SCMA selects a single memory state and directly copies it to the Copy-LSTM which includes an adaptive copy mechanism in its internal structure and can choose weather to "copy" or "edit" an existing word. The textual attended vector is gated to remove incorrect attended words before being passed as an input to the Copy-LSTM along with the visual attention vector. EditNet is equiped with an LSTM-based denoising auto-encoder (right) which takes as input the existing caption, encodes it into a compressed representation and then decodes the compressed representation to its expected output. }
    \label{model}
\end{figure*}

\section{Related Work}

\subsection{Image Captioning}

Image captioning has been widely studied in the computer vision and natural language processing communities for multiple decades. Traditional captioning systems, which were primarily used for video captioning, involved detecting keywords and using these keywords to fill in hand-made templates \cite{pan2004gcap,pan2004automatic}. These models had the advantage of always producing logical sentence structures, but their expressive power was severely limited due to the need for researchers to manually design templates. 

In the past five years, neural network-based image captioning models have risen to prominence. Introduced by \cite{vinyals2015show}, these approaches fall into the broader category of encoder-decoder models, alongside those for machine translation, summarization, speech recognition, and a host of other tasks \cite{sutskever2014sequence}. Specifically, \cite{vinyals2015show} proposed a captioning model consisting of a CNN encoder and an LSTM decoder, in which the output of the CNN encoder (the final convolutional layer) was used as input to the LSTM. \cite{Xu2015ShowAA} dramatically improved upon the model introduced by \cite{vinyals2015show} with the addition of an attention mechanism. These mechanisms have engendered large performance improvements across sequence learning tasks \cite{Xu2015ShowAA,bahdanau2014neural,chorowski2015attention}.

Of the attention mechanisms designed specifically for image captioning, bottom-up and top-down attention (Up-Down; \cite{Anderson2017BottomUpAT}) and the recent attention-on-attention (AoA; \cite{Huang2019AttentionOA}) have proven among the most successful. \cite{Anderson2017BottomUpAT} adds a top-down attention LSTM before the language LSTM to selectively attend to spatial image features. \cite{Huang2019AttentionOA}, currently the state-of-the-art, adds an attention-on-attention module after both the language LSTM and the standard attention mechanism. This module is designed to measure the relevance between the attention result and the query; it transforms the output of the standard attention mechanism, multiplying it element-wise by an attention gate (a different transformation of the output followed by a sigmoid function). 

Finally, parallel to improvements in attention mechanisms, \cite{Rennie2016SelfCriticalST} proposed a new optimization objective for image captioning. Traditionally, image captioning models are trained to minimize the cross-entropy between their word-level output and the ground truth caption. \cite{Rennie2016SelfCriticalST} instead optimizes a sequence-level objective, such as CIDEr \cite{Vedantam2014CIDErCI} or METEOR \cite{Banerjee2005METEORAA}, using methods adopted from reinforcement learning. It is now common in the literature to evaluate the performance of new models using both cross-entropy and self-critical training. 

\subsection{Sequence-to-Sequence Editing}

In the past year, a new paradigm based around editing the output of sequence-to-sequence models has been shown to improve the performance of a large class of models. \cite{Hashimoto2018ARF} proposed a retrieve-and-edit framework for generating sequences such as source code from natural language inputs. The authors showed strong performance gains over standard sequence-to-sequence models on a code autocompletion task and the Hearthstone cards benchmark; these results suggests that editing existing outputs may be easier than generating new outputs from scratch. 

Most recently, \cite{Sammani2019LookAM} proposed the task of editing image captions.  \cite{Sammani2019LookAM} used a deep averaging network to encode an existing caption (outputted by a traditional sequence-to-sequence model) into a vector, and added a gated output of this vector along with the LSTM output, enabling the LSTM to model the ``residual'' information. This model improved upon the performance of some baseline methods, but failed to outperform the state-of-the-art in image captioning. 

In this work, we introduce a new framework for the editing task proposed in \cite{Sammani2019LookAM}, where we employ copy mechanisms to better take advantage of the information in the existing caption. Compared to both \cite{Sammani2019LookAM} and the state-of-the-art in image caption generation (discussed above) \cite{Huang2019AttentionOA}, we show significant performance improvements across image captioning metrics. 

\section{Proposed Methodology}
Our model consists of two submodules: \textit{EditNet} and \textit{DCNet}. In the following sections, we describe each of these submodules in detail. A complete overview of our model is shown in Figure \ref{model}.

\subsection{EditNet}
EditNet is a model designed to learn whether to copy or edit each word in an input caption. It has an encoder-decoder structure with two components: (1) a Selective Memory Attention Mechanism (SCMA) and (2) a Copy-LSTM decoder. We describe these parts in detail in the following subsections. 

\subsubsection{EditNet Architecture}

We base the general structure of EditNet on the widely-used bottom-up and top-down architecture from \cite{Anderson2017BottomUpAT}. 

Given an image, our encoder extracts a set of 36 visual features using an R-CNN based network. We denote these features by  $V = \left\{v_{1}, v_{2},\ldots, v_{k}\right\}$ where $v_{i}\in \mathbb{R}^{2048}$ and \textit{k} is the number of objects (in our case, $k=36$). 

Given output of our encoder and an input caption, our decoder produces an edited version of the input caption. Like \cite{Anderson2017BottomUpAT}, our decoder contains an attention LSTM and a language LSTM. Unlike previous work, we add an input caption LSTM and a novel SCMA module, and we replace the language LSTM with a Copy-LSTM.

First, we encode the input caption using a uni-directional one-layer LSTM (see the dashed red box in Figure \ref{model}). In the following sections, we denote the encoded input caption by $\overline{h}_{s}=\left[h_{1}^{e} \ldots h_{n}^{e}\right]$, where $n$ is the number of words in the input caption. We denote the memory states of the corresponding LSTM cells by $\overline{c}_{s}=\left[c_{1}^{e} \ldots c_{n}^{e}\right]$. 

Next, we feed the following inputs to the attention LSTM: the word embedding vector, the last hidden state of the caption encoder, the mean-pooled image features $\overline{\boldsymbol{v}}=\frac{1}{k} \sum_{i} \boldsymbol{v}_{i}$, and the previous hidden state of the language LSTM. That is, we input $x_{t}^{1}=\left[w_{t} ; h_{n}^{e} ; \bar{v} ; h_{t-1}^{2}\right]$ where $;$ indicates concatenation. Note that this attention LSTM is a standard LSTM, not a Copy-LSTM, because it does not take input from the SCMA module. The output of the attention LSTM $h_{t}^{1}$ is used to compute two attention vectors, one over the visual features and another over the textual features. These are fused with a gating mechanism and used as input to the Copy-LSTM. 


The attention weights over textual features are also used as input to the SCMA module; this module may be thought of as learning to select and copy from the input caption LSTM, and its output is used as input to the Copy-LSTM. The Copy-LSTM takes as input the outputs of the attention LSTM along with the visual attended vector and the gated textual vector. It outputs a hidden state $h_{t}^{2}$, which is passed to a final linear layer to predict the softmax probability distribution over the vocabulary. Finally, this distribution is fused with the output of the denoising autoencoder (described in section 3.2) to produce the final output word.

\subsubsection{Selective Copy Memory Attention (SCMA)}

\begin{figure}
    \centering
    \includegraphics[width=0.4\textwidth]{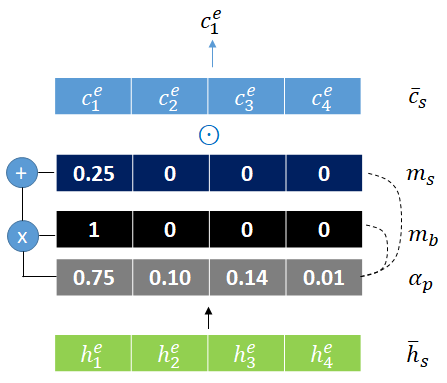}
    \caption{The operational flow of SCMA. Attention weights (grey) are computed from the encoded output of the input caption and highlight the importance of each word in accordance to the current generated word at the language model. The attention weights are then used to calculate two masks: a binary mask $m_{b}$ which is multiplied with the corresponding attention weight of each word, and a shifting mask $m_{s}$ which shifts the multiplication result to 1. Finally, each resulting element is multiplied with the corresponding memory state. Eventually, all memory states are eliminated except for the one with the maximum attention weight, which is the final copied output.}
    \label{scma}
\end{figure}

The SCMA (Figure \ref{scma}) enables our model to select and copy memory states corresponding to words in the input caption. 

We measure the similarity between the current initial decoder output $h_{t}^{1}$ and each word in the previous caption $\overline{h}_{s}$ using a shallow neural network followed by a softmax:
\begin{equation}
\alpha_{p}=\operatorname{softmax}\left(w_{a}^{T} \tanh \left(W_{s} \bar{h}_{s}+W_{h} h_{t}^{1}\right)\right)
\end{equation}

Different from the conventional attention mechanism, we do not utilize $\alpha_{p}$ directly. Instead, we utilize the corresponding memory state in the input caption encoder LSTM. To be precise, we copy the corresponding memory state $c_{t}^{e}$ from the input caption encoder with the highest similarity (i.e. highest softmax output from $\alpha_{p}$). 

Notably, this indexing operation is non-differentiable. To get around this problem, we employ the re-parametrization trick \cite{Kingma2013AutoEncodingVB}. We construct two masks, a binary mask and a shifting mask. The binary mask $m_{b}$ includes a 1 in the index of the maximum probability value of the softmax output $\alpha_{p}$, while the shifting mask $m_{s}$ includes the residual values that shift the result of $\alpha_{p} m_{b}$ of the maximum word to 1 and 0 otherwise. Mathematically, this operation is:
\begin{equation}
c_{S}^{e}=\sum_{i=1}^{n}\left(\alpha_{p_{i}} m_{b_{i}}+m_{s_{i}}\right) c_{i}^{e}
\end{equation}

For example, if the attention weight for the maximum word is 0.8, then $m_{b_{i}}$ = 1 and $m_{s_{i}}$ = 0.2. Therefore, the extracted memory state would be $c_{i}^{e}(0.8 \cdot 1 + 0.2) = c_{i}^{e}$. Similarly, if the attention weight for a non-maximum word is 0.3, then $m_{b_{i}}$ = 0 and $m_{s_{i}}$ = 0. In this case, $c_{i}^{c}(0.3 \cdot 0+0)=0$, and $c_{i}^{e}$ would be eliminated. Consequently, all words with a probability lower than the maximum value would be multiplied by $0$, and the memory cell $c_{i}^{e}$ with the maximum probability would remain. 

We utilize the copied memory state, denoted $C_{S}^{e}$, in the Copy-LSTM described below. 

\begin{figure}
    \centering
    \includegraphics[width=0.4\textwidth]{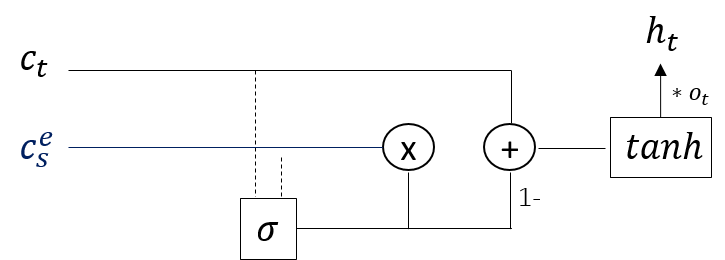}
    \caption{The structure of our Copy-LSTM (Equations 4-6).}
    \label{copylstm}
\end{figure}

\subsubsection{Copy-LSTM}
To incorporate the information from the input caption and SCMA module into the language decoder, we augment the LSTM cell with an adaptive copy mechanism. Our modified LSTM cell, which we denote Copy-LSTM, includes a ``copy gate'' that controls how much information is taken from the SCMA module relative to the other input sources (the visual features and the hidden state). 

We now give a mathematical formulation for the Copy-LSTM. As in a standard LSTM, the forget gate, input gate, output gate and memory state calculation are given as:
\begin{equation}
\begin{aligned} f_{t} &=\sigma\left(W_{f} \cdot\left[h_{t-1}, x_{t}\right]+b_{f}\right) \\ i_{t} &=\sigma\left(W_{i} \cdot\left[h_{t-1}, x_{t}\right]+b_{i}\right) \\ \tilde{C}_{t} &=\tanh \left(W_{C} \cdot\left[h_{t-1}, x_{t}\right]+b_{C}\right) \\ C_{t} &=f_{t} * C_{t-1}+i_{t} * \tilde{C}_{t} \\ o_{t} &=\sigma\left(W_{o}\left[h_{t-1}, x_{t}\right]+b_{o}\right) \end{aligned}
\end{equation}

In addition, we incorporate a copy gate, $c_{g_t}$, which may be thought of as calculating the similarity between the copied memory state and the word currently being generated:
\begin{equation}
c_{g_{t}}=\sigma\left(W_{n} \cdot\left[C_{t}, C_{s}^{e}\right]\right)
\end{equation}

We then compute the amount to take from both memory states, and modify the LSTM memory state to: 
\begin{equation}
C_{a p_{t}}=c_{g_{t}} * C_{s}^{e}+\left(1-c_{g_{t}}\right) * C_{t}
\end{equation}

The hidden state is then computed with a tanh activation function of the newly constructed memory state, multiplied by the output gate:
\begin{equation}
h_{t}=o_{t} * \tanh \left(C_{a p_{t}}\right)
\end{equation}

With these modifications, the Copy-LSTM is able to incorporate the desired information into its output representation $h_{t}$. It passes this hidden state to the output layer, which predicts the next word in the caption. Note that if the gate $c_{g_t}$ is $1$, then the word from the input caption is fully copied, and if it is $0$, then the previous caption is ignored and the word is generated anew. An overview of the modified internal structure of the LSTM (Copy-LSTM) is shown in Figure \ref{copylstm}.


\subsubsection{Context Gating}
As mentioned earlier, our model attends over the textual features $\bar{h}_{s}$ of the existing caption. Intuitively, however, attending over the textual features may mislead the language LSTM when the existing caption contains incorrect information. Inspired by the recent advances in neural machine translation, we incorporate a ``context gate" that learns how much to focus on the source context (the textual attended feature vector) and the target context (the word embedding vector and the current LSTM hidden state). That is:
\begin{equation}
\left.c_{m}=z_{t} \odot \tanh \left(W_{s} c_{t}\right)+\left(1-z_{t}\right) \odot \tanh \left(W_{t} \cdot\left[w_{t} ; h_{t}\right]\right)\right)
\end{equation}

where
\begin{equation}
z_{t}=\sigma\left(W_{Z} \cdot\left[w_{t} ; h_{t} ; c_{t}\right]\right)
\end{equation}

and
\begin{equation}
c_{t}=\sum_{i=1}^{n} \alpha_{p_{i}} h_{i}^{e}
\end{equation}

Note that equation 7 is different from \cite{Tu2016ContextGF}, where we include the gate and its compliment before the activation function. We find that this operation performs better in completely removing the unwanted elements in the attention vector. Also note that $\alpha_{p}$ in equation 9 is same as that of equation 1. We find that sharing the parameters give better similarity scores and reduces the number of overall parameters. 

\subsection{Denoising Captioner (DCNet)}

In parallel to EditNet, we use a denoising auto-encoder, denoted DCNet, to denoise our input caption. Denoising auto-encoders are traditionally used to re-construct noisy images. In our case, we may think of our input caption as a noisy version of a true caption. 

DCNet is composed of a bi-directional LSTM encoder, which encodes the noisy caption into a compressed representation, and an LSTM decoder, which decodes the compressed representation. Note that DCNet operates only on textual features, without any visual features. Additionally, the parameters in DCNet are not shared with the parameters in EditNet. Further details on DCNet are included in the supplementary material. 


\subsection{Objectives}

We first train our model by optimizing the cross entropy (XE) loss:

\begin{equation}
L_{X E}(\theta)=-\sum_{t=1}^{T} \log \left(p_{\theta}\left(y_{t}^{*} | y_{1: t-1}^{*}\right)\right)
\end{equation}

After training with cross-entropy, we additionally optimize our language decoder using mean-squared error between the last decoder hidden state of the language model and the last hidden state of the ground truth caption. This ground truth caption hidden state is obtained by running the ground-truth caption through the encoder of the denoising auto-encoder. In sum, this loss is given by: 

\begin{equation}
L_{M S E}=\frac{1}{n} \sum_{i=1}^{n}\left(h_{n}^{d}-h_{n}^{g}\right)
\end{equation}

\noindent where we linearly project the last hidden state of the language model $h_{n}^{2}$ without using any activation function:

\begin{equation}
h_{n}^{d}=W_{d} h_{n}^2+b_{d}
\end{equation}

We provide ablation studies on this scheme in section 4.4, where we demonstrate an increase in the CIDER score of DCNet from 1.171 to 1.183. This optimization scheme is simple and not restricted to our model; it can be applied to general sequence-to-sequence or vector-to-sequence task. Our final loss (for non-sequence-level training) is:
\begin{equation}
L=L_{X E}(\theta)+L_{M S E}
\end{equation}
 
Finally, for comparison with other works, we directly optimize for CiDEr-D using sequence-level training \cite{Rennie2016SelfCriticalST}. As in \cite{Rennie2016SelfCriticalST}, the policy gradient is:
\begin{equation}
\nabla_{\theta} L_{R L}(\theta) \approx-\left(r\left(C^{s}\right)-b\right) \nabla_{\theta} \log p_{\theta}\left(C^{s}\right)
\end{equation}

\noindent where $r\left(C_{i}^{s}\right)$ is the CIDEr score of the sampled caption and $b$ is the CIDEr score of a greedily decoded caption \cite{Rennie2016SelfCriticalST}.
\begin{table*}[]
\begin{center}
\renewcommand\arraystretch{1.15}
\caption{Performance of our \textbf{single model} and other state-of-the-art models on MS-COCO “Karpathy” test split under cross-entropy training.  All values are reported as percentage (\%). * indicates the results obtained from the publicly available pre-trained model. - indicates that the results are not provided. $^\dagger$ indicates results from previous models trained to edit captions, rather than generate captions. }
\vspace{1mm}

\begin{tabular}{|l|ccccccc|} 
\hline
\multicolumn{1}{|c|}{\textbf{Model}}  & \multicolumn{7}{c|}{Cross-Entropy Loss}                                                            \\ \hline
\multicolumn{1}{|c|}{\textbf{Metric}} & BLEU-1                             & \multicolumn{1}{l}{BLEU-2}         & \multicolumn{1}{l}{BLEU-3}         & \multicolumn{1}{l}{BLEU-4}         & \multicolumn{1}{l}{ROUGE-L}        & \multicolumn{1}{l}{CIDEr-D}         & \multicolumn{1}{l|}{SPICE} \\ \hline
NIC\cite{vinyals2015show}                                   & -                                  & -                                  & -                                  & 29.6                               & 52.6                               & 94.0                                & -                         \\ \cline{1-1}
SCST  \cite{Rennie2016SelfCriticalST}                                & -                                  & -                                  & -                                  & 30.0                               & 53.4                               & 99.4                                &             -              \\ \cline{1-1}
Adaptive  \cite{Lu2016KnowingWT}                            & 74.2                               & 58.0                               & 43.9                               & 33.2                               & 54.9                               & 108.5                               & 19.4                      \\ \cline{1-1}
Up-Down   \cite{Anderson2017BottomUpAT}                            & 77.2                               & -                                  & -                                  & 36.2                               & 56.4                               & 113.5                               & 20.3                      \\ \cline{1-1}

MN (Up-Down)$^{\dagger}$       \cite{Sammani2019LookAM}                          & 76.9                                & 61.2                               & 47.3                               & 36.1                               & 56.4                         & 112.3                     & 20.3                            \\ \cline{1-1}
RFNet       \cite{Jiang2018RecurrentFN}                          & 76.4                               & 60.4                               & 46.6                               & 35.8                               & 56.8                               & 112.5                               & 20.5                      \\ \cline{1-1}
GCN-LSTM      \cite{Yao2018ExploringVR}                        & 77.3                               & -                                  & -                                  & 36.8                               & 57.0                               & 116.3                               & 20.9                      \\ \cline{1-1}

AoANet         \cite{Huang2019AttentionOA}                       & 77.4                               & -                                  & -                                  & 37.2                               & 57.5                               & 119.8                               & \textbf{21.3}             \\\cline{1-1}

AoANet*       \cite{Huang2019AttentionOA}                        & 77.3                               & 61.6                               & 47.9                               & 36.9                               & 57.3                               & 118.4                               & \textbf{21.6}            \\ \hline

\textbf{ETN (Ours) }                          & \textbf{77.9} & \textbf{62.5} & \textbf{48.9} & \textbf{38.0} &\textbf{57.7} & \textbf{1.200} & 21.2 \\ \hline
\end{tabular}

\end{center}
\end{table*}

\begin{table*}[]
\begin{center}
\renewcommand\arraystretch{1.15}
\caption{Performance of our \textbf{single model} and other state-of-the-art models on MS-COCO “Karpathy” test split under CIDER-D score optimization.  All values are reported as percentage (\%). * indicates the results obtained from the publicly available pre-trained model. - indicates that the results are not provided.}

\vspace{1mm}

\begin{tabular}{|l|ccccccc|}
\hline
\multicolumn{1}{|c|}{\textbf{Model}}  & \multicolumn{7}{c|}{Sequence-Level Optimization}                                                                                                                                                                                                              \\ \hline
\multicolumn{1}{|c|}{\textbf{Metric}} & BLEU-1                             & \multicolumn{1}{l}{BLEU-2}         & \multicolumn{1}{l}{BLEU-3}         & \multicolumn{1}{l}{BLEU-4}         & \multicolumn{1}{l}{ROUGE-L}        & \multicolumn{1}{l}{CIDEr-D}         & \multicolumn{1}{l|}{SPICE} \\ \hline
NIC \cite{vinyals2015show}                         & -                                  & -                                  & -                                  & 31.9                               & 54.3                               & 106.3                               & -                          \\ \cline{1-1}
SCST    \cite{Rennie2016SelfCriticalST}                     & -                                  & -                                  & -                                  & 34.2                               & 55.7                               & 114.0                               & -                          \\ \cline{1-1}
Up-Down       \cite{Anderson2017BottomUpAT}               & 79.8                               & -                                  & -                                  & 36.3                               & 56.9                               & 120.1                               & 21.4                       \\ \cline{1-1}
RFNet        \cite{Jiang2018RecurrentFN}                & 79.1                               &        63.1                            &              48.4                      & 36.5                               & 57.3                               & 121.9                               & 21.2                       \\ \cline{1-1}
GCN-LSTM        \cite{Yao2018ExploringVR}             & 80.5                               &          -                          &                                  & 38.2                               & 58.3                               & 127.9                               & 22.0                       \\ \cline{1-1}

AoANet*       \cite{Huang2019AttentionOA}               & 80.5                               & 65.2                               & 50.1                               & 39.1                               & 58.9                               & \textbf{128.9}                      & 22.7                       \\ \cline{1-1}

AoANet     \cite{Huang2019AttentionOA}                  & 80.2                               &                 -                   &         -                           & 38.9                               & 58.8                               & \textbf{129.8}                               & 22.4                       \\ \hline

\textbf{ETN (Ours)}          & \textbf{80.6} & \textbf{65.3} & \textbf{51.1} & \textbf{39.2} & \textbf{58.9} & \textbf{128.9} & \textbf{22.6}              \\ \hline
\end{tabular}
\end{center}
\end{table*}

\subsection{Implementation Details}
\par{\textbf{EditNet}}: 
For visual features, we use bottom-up features from \cite{Anderson2017BottomUpAT}. For textual features, we use captions from \cite{Huang2019AttentionOA}.\footnote{We use the pretrained model: https://github.com/husthuaan/AoANet}

We set the embedding and hidden size of both the LSTM encoder and decoder network to 1024 and the attention dimension to 512. We train EditNet for 15 epochs using cross-entropy, as in equation 12. Note that for EditNet, we do not use MSE optimization after training with cross-entropy. However, we still provide ablation studies on training EditNet with MSE optimization. 

We use the ADAM optimizer \cite{Kingma2014AdamAM} with batch size 80,  initial learning rate 5e-4, and decay the learning rate decay by a factor of 0.8 every 3 epochs. We increase the scheduled sampling probability by 0.05 every 5 epochs \cite{Bengio2015ScheduledSF}. We optimize the CIDEr-D score with sequence-level training for another 25 epochs with an initial learning rate of 5e-5 and anneal it by 0.5 when the CIDER-D score shows no improvement for one epoch. We do not use label smoothing. 

\par{\textbf{DCNet}}: DCNet fully operates on textual features, without using any visual features. For the encoder LSTM, we set the hidden size to 512 for each direction, ending up with a dimension of 1024 for both directions. For the decoder, we choose the top-down decoder \cite{Anderson2017BottomUpAT} with a hidden size of 1024. The embedding dimension is set to 1024 and the attention dimension to 512. We train DCNet for 4 epochs using cross-entropy, as in equation 10 and optimize it with MSE for one additional epoch, as in equation 13. We set the batch size to 60 and use the same optimization settings (for both XE and CIDER-D optimization) as EditNet.

\section{Experiments}

\begin{figure*}
    \centering
    \includegraphics[width=0.9\textwidth,height=\textheight,keepaspectratio]{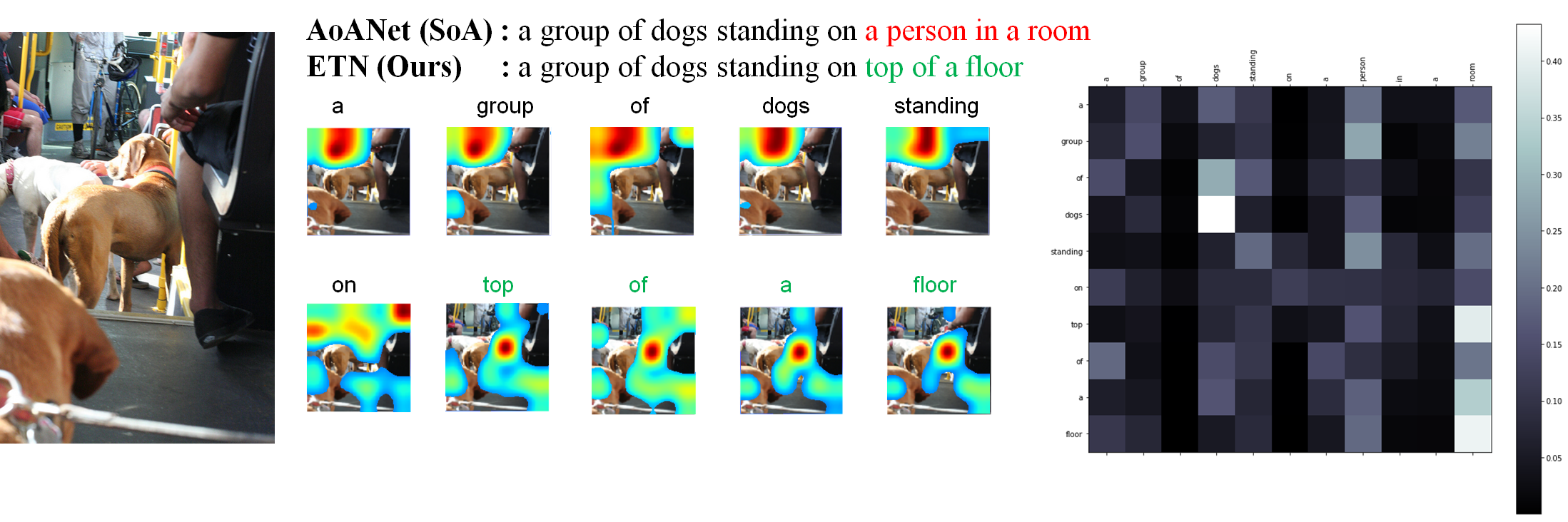}
    \caption{A caption generated by our editing framework when supplied with an input captions from AoANet \cite{Huang2019AttentionOA}, along with its visual attention maps (left) and textual alignment plots (right). The alignment plot visualizes the words selected by the SCMA mechanism and copied to the Copy-LSTM.}    \label{results}
\end{figure*}

\begin{figure}
    \centering
    \includegraphics[width=0.5\textwidth]{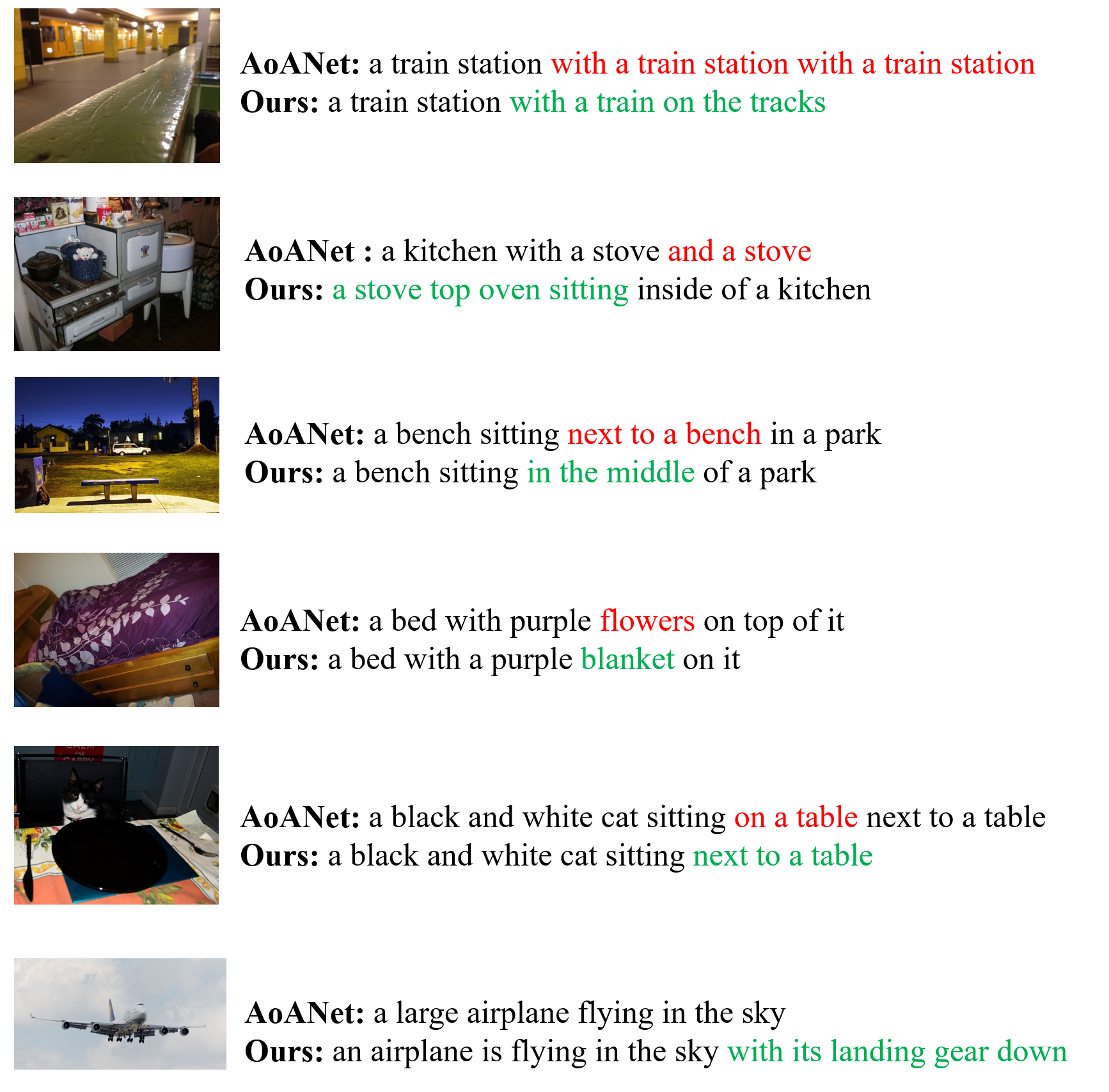}
    \caption{More results from our model compared to AoANet \cite{Huang2019AttentionOA}}. 
    \label{moreres}
\end{figure}

\subsection{Dataset}
We evaluate our proposed method on the popular MS COCO dataset \cite{Lin2014MicrosoftCC}, which contains 123,287 images labeled with 5 captions for each by 5 different people. We use the standard “Karpathy” data split \cite{Karpathy2015DeepVA} for the offline performance comparisons, in which 5,000 images are used for validation, 5,000 are used for testing and 113,287 are used for training. We convert all sentences to lower case and remove words that occur fewer than 3 times from our vocabulary, ending up with a vocabulary of 13,368 words. For evaluation, we use 4 different metrics: BLEU (1- to 4-grams) \cite{Papineni2001BleuAM}, ROUGE-L \cite{Lin2004ROUGEAP}, CIDEr-D \cite{Vedantam2014CIDErCI} and SPICE \cite{Anderson2016SPICESP}. All metrics are computed with the standard public evaluation code. \footnote{https://github.com/tylin/coco-caption}

\subsection{Quantitative Analysis}

\par{\textbf{Offline Evaluation:}} We report the performance of our model compared with the current state-other-art in Tables 1 and 2. These models include NIC \cite{vinyals2015show}, which uses a vanilla CNN-LSTM encoder-decoder framework; SCST \cite{Rennie2016SelfCriticalST}, which optimizes a standard attention-based model using non-differentiable metrics; Adaptive \cite{Lu2016KnowingWT} which uses a visual sentinel to eliminate visual attention over non-visual words; Up-Down \cite{Anderson2017BottomUpAT} which uses an attention LSTM to attend over image features extracted from a Faster R-CNN model; RFNet \cite{Jiang2018RecurrentFN} which uses multiple CNNs and LSTMs that are connected to each other; GCN-LSTM \cite{Yao2018ExploringVR} which predicts an image scene graph and fuses it with the visual features to produces better feature vectors; and finally AoANet \cite{Huang2019AttentionOA} which uses a Transformer-based language model and filters incorrect elements out of the attended visual vector. 

For the cross-entropy loss training stage in Table 1, our single model achieves the highest score on all metrics with the exception of SPICE, where its score is marginally lower than \cite{Huang2019AttentionOA}. For the sequence-level optimization stage, our model also achieves the highest scores across all metrics except CIDER-D, where is slightly below the published results from \cite{Huang2019AttentionOA} and equal to the pretrained model released by \cite{Huang2019AttentionOA}. Our model also dramatically outperforms the only other caption-editing model, Modification Networks (MN) \cite{Sammani2019LookAM}, when using cross-entropy training (sequence-level results for MN are not reported). 

\par{\textbf{Online Evaluation:}} The performance of our model on the official MS-COCO online testing server is provided in the supplementary material.  
\subsection{Qualitative Analysis}
Figure \ref{results} and \ref{moreres} show some results generated by our editing framework. In Figure \ref{results}, we can see that the current state-of-art framework \cite{Huang2019AttentionOA} generates a sentence where it recognizes the correct objects, but fails to arrange them in the correct format (standing \textit{on a person}). Using these captions as input to our editing framework, our model is able to fix the sentence and arrange the objects in the correct format (standing \textit{on top of a floor}). The right side of Figure \ref{results} shows a visualization of the textual alignment between the detected words (y-axis) and the existing words (x-axis). From this, we can see which words the SCMA mechanism selected and copied to the Copy-LSTM. 

Figure \ref{moreres} demonstrates that our model is also capable of replacing repetitive words and adding details to captions. The first three examples show that AoANet often repeats words when it is unable to recognize the correct details in the image (e.g. \textit{with a train station}, \textit{and a stove}, \textit{a bench}). Our editing model successfully fixes these issues by replacing the repetitive words. The last example in Figure \ref{moreres} demonstrates that our model can add additional details to an existing caption, even when the visual features are minimal in the image (\textit{with its landing gear down}).

\subsection{Ablation Studies}
In this section, we provide ablation studies on using mean-squared error (MSE) optimization after training with cross-entropy, and on using the Copy-LSTM alone along with the SCMA mechanism. 

Table 3 gives results for EditNet and DCNet with and without using MSE optimization. The results without MSE are obtained after training each submodule with cross-entropy (XE) loss, while the results with MSE optimization are after optimizing EditNet with both XE (first alone) and MSE (together with XE). For DCNet, the addition of one epoch of MSE training boosts the BLEU-4 score from 36.9 to 37.2 and the CIDEr-D score from 117.1 to 118.3. We also examine the performance of our Copy-LSTM alone: we remove the visual features, the context gate and the DCNet sub-module, and we train the EditNet with cross-entropy (without MSE optimization). Our scores for  BLEU-1/BLEU-2/BLEU-3/BLEU-4/ROUGE-L/CIDER-D are \textbf{77.3, 61.7, 48.0, 37.0, 57.2} and \textbf{117.3}, respectively. This translates to no improvement over the pre-trained AoANet model for some metrics and a very small improvement for others. Moreover, we examine the performance of the context gate in our EditNet sub-module without any visual features, and find that the context gate improves the CIDER-D score from 117.3 to 117.5. By contrast, EditNet with visual features and textual context gating gives better scores across the board. Finally, we examine the performance of the non-differentiable indexing in the SCMA mechanism. We find that simply using soft-attention on the memory states achieves a CIDER-D score of 119.2, which is lower than the achieved score of 1.200 when using non-differentiable indexing.

\begin{table}[]
\caption{The effect of using MSE optimization after cross-entropy training. B-4 indicates BLEU-4 and C indicates CIDEr-D.} 
\begin{tabular}{|c|c|c|c|c|}
\hline
\textbf{Model}       & \multicolumn{2}{c|}{\textbf{DCNet}} & \multicolumn{2}{c|}{\textbf{EditNet}} \\ \hline
\textbf{Metrics}     & \textbf{B-4}      & \textbf{C}      & \textbf{B-4}       & \textbf{C}       \\ \hline
w/o MSE Optimization & 36.9              & 117.1           & 38.0               & 118.0            \\ \hline
w/ MSE Optimization  & 37.2              & 118.3           & 38.0               & 118.5            \\ \hline
\end{tabular}
\end{table}

\section{Conclusion}
In this paper, we propose a framework for editing image captions based on iterative adaptive refinement of an existing caption. This new perspective enables our framework to focus on fixing details of existing captions, rather than generating new captions from scratch. Specifically, our model consists of two novel sub-modules: (1) \textit{EditNet}, a language module with an adaptive copy mechanism (\textit{Copy-LSTM}) and a Selective Copy Memory Attention mechanism (\textit{SCMA}), and (2) \textit{DCNet}, an LSTM-based denoising auto-encoder. Experiments on the MS COCO dataset demonstrate that our approach achieves state-of-art performance across image captioning metrics. In the future, our framework may be extended to related tasks such as visual question answering and neural machine translation. 

{\small
\bibliographystyle{ieee_fullname}
\bibliography{etn_final}
}

\end{document}